\def\year{2021}\relax
\documentclass[letterpaper]{article} 
\usepackage{aaai21}  
\usepackage{times}  
\usepackage{helvet} 
\usepackage{courier}  
\usepackage[hyphens]{url}  
\usepackage{graphicx} 
\usepackage{natbib}  
\usepackage{caption} 
\usepackage[switch]{lineno} 
\usepackage{amssymb}
\usepackage{amsmath}
\usepackage{mathtools}
\usepackage{breqn}
\usepackage{subcaption}
\usepackage[T1]{fontenc}
\usepackage{listings}
\usepackage{xspace}
\usepackage{paralist}
\usepackage{multirow}
\usepackage{listings}
\usepackage{xcolor}

\usepackage{rotating}

\newcommand{\checknew}[1]{{\color{black}{#1}}}

\usepackage{url}
\newcommand{\eg}{\mbox{\it e.g.}}
\newcommand{\ie}{\mbox{\it i.e.}}
\newcommand{\metaDecisionSpace}{\mathcal{M}}
\newcommand{\accept}{\texttt{\footnotesize{Accept}}\xspace}
\newcommand{\override}{\texttt{\footnotesize{Solve}}\xspace}
\newcommand{\acceptS}{\texttt{A}\xspace}
\newcommand{\overrideS}{\texttt{S}\xspace}

\newcommand{\x}{x}
\newcommand{\decisionMaker}{d}
\newcommand{\labelSpace}{\mathcal{Y}}
\newcommand{\classifier}{h}

\newcommand{\utilityFunc}{U}

\newcommand{\metaDecisionFunc}{m}

\newcommand{\correctLabel}{y}
\newcommand{\predictedLabel}{\hat{y}}
\newcommand{\recommendation}{r}
\newcommand{\mistakePenalty}{\beta}
\newcommand{\overrideCost}{\lambda}
\newcommand{\humanAccuracy}{a}
\newcommand{\threshold}{c(\mistakePenalty, \overrideCost, \humanAccuracy)}
\newcommand{\expect}{\mathbb{E}}
\newcommand{\euFunc}{\psi}

\newcommand{\prob}{P}

\DeclareMathOperator*{\argmax}{arg\,max}
\let\oldcdot=\cdot
\def\cdot{\negthinspace\oldcdot\negthinspace}

\newcommand{\toy}{Scenario1\xspace}

\usepackage[flushleft]{threeparttable}

\newcommand{\abbrvlossname}{{\em team-loss}\xspace}
\newcommand{\logloss}{log-loss\xspace}

\title{Is the Most Accurate AI the Best Teammate? Optimizing AI for Teamwork}
\author{
Gagan Bansal$^1$\hspace{10pt}
Besmira Nushi$^2$\hspace{10pt}
Ece Kamar$^2$\hspace{10pt}
Eric Horvitz$^2$\hspace{10pt}
Daniel S. Weld$^{1,3}$\\
{\normalfont
$^1$University of Washington\hspace{10pt}
$^2$Microsoft Research\hspace{10pt}
$^3$Allen Institute for AI
}
}
\begin{document}
\maketitle

\begin{abstract}
AI practitioners typically strive to develop the most {\em accurate} systems, making an implicit assumption that the AI system will function autonomously. However, in practice, AI systems often are used to provide {\em advice} to people in domains ranging from criminal justice and finance to healthcare. In such {\em AI-advised decision making}, humans and machines form a {\em team}, where the human is responsible for making final decisions.   But is the most accurate AI the best teammate? We argue ``No'' --- predictable performance may be worth a slight sacrifice in AI accuracy.
Instead, we argue that AI systems should be trained in a human-centered manner, directly optimized for {\em team performance}. We study this proposal for a specific type of human-AI teaming, where the human overseer chooses to either {\em accept} the AI recommendation or {\em solve} the task themselves. To optimize the team performance for this setting we maximize the team's {\em expected utility}, expressed in terms of the quality of the final decision, cost of verifying, and individual accuracies of people and machines. Our experiments with linear and non-linear models on real-world, high-stakes datasets show that the most accuracy AI may not lead to highest team performance and show the benefit of modeling teamwork during training through improvements in expected team utility across datasets, considering parameters such as human skill and the cost of mistakes. We discuss the shortcoming of current optimization approaches beyond well-studied loss functions such as \logloss, and encourage future work on AI optimization problems motivated by human-AI collaboration.
\end{abstract}

\section{Introduction}

Many AI systems are developed for use in collaborative settings, where people work with an AI teammate. For example, numerous applications of AI have been designed as advisory tools, providing input to people who are tasked with making final decisions. Beyond the appropriateness of people making the final calls, the advisory role of AI systems may be obligatory; legal requirements may prohibit complete automation~\cite{gdpr,face-recognition-law}. Studies have demonstrated domains and tasks where human-AI teams may perform better than either the AI or human alone~\cite{nagar2011making,patel2019human,kamar2012combining}. For human-AI teams, optimizing the performance of the whole team is more important than optimizing the performance of an individual member. 
Yet, to date, the AI community has focused on maximizing the individual accuracy of machine-learned models, assuming implicitly that this will optimize team performance. This raises an important question: Is the most accurate AI the best possible teammate for a human? 

\begin{figure}[t]
    \centering
    \includegraphics[width=\linewidth]{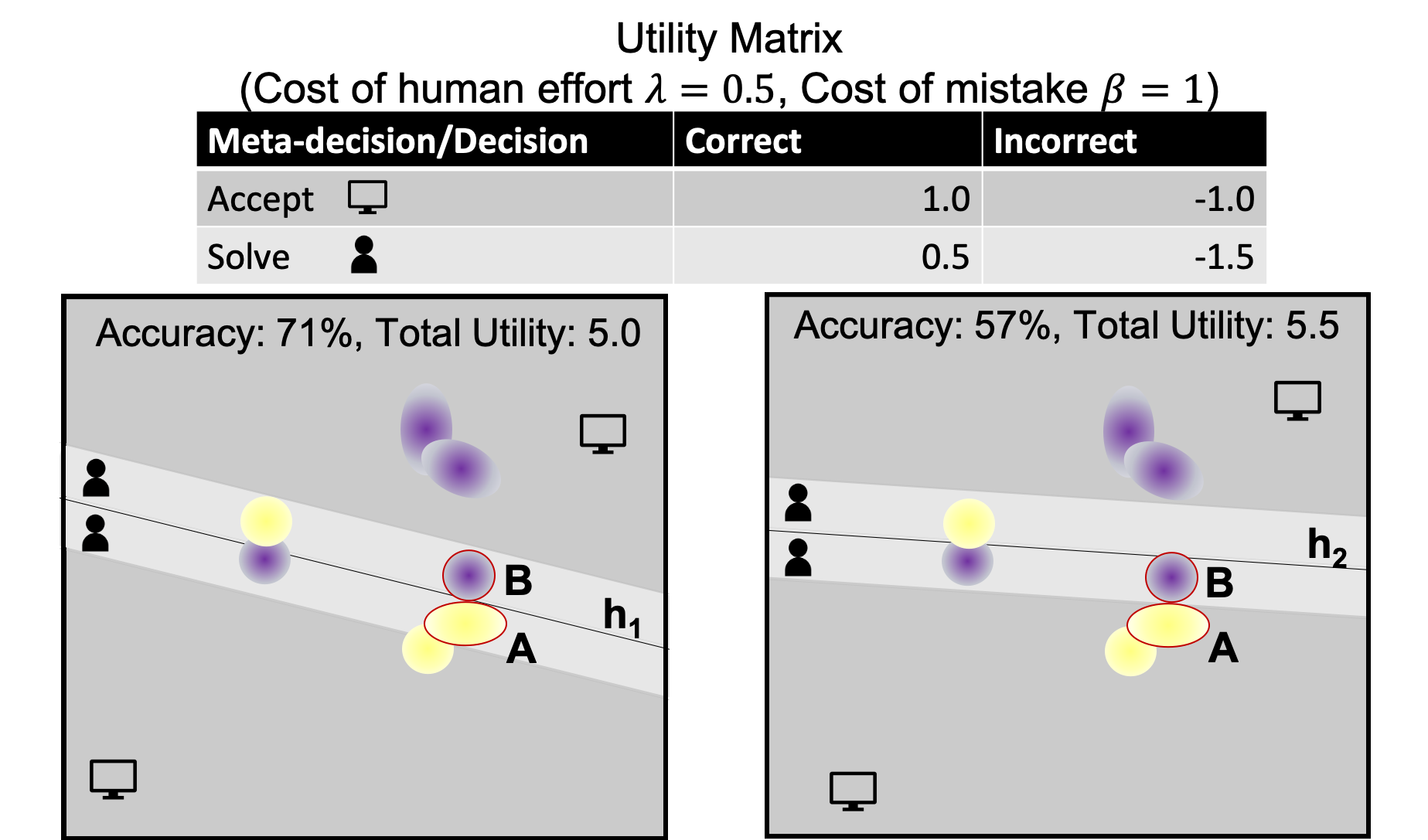}
    \caption{Consider a binary classification problem (purple vs. yellow). Assume each blob is uniformly distributed and of the same size. In a human-AI team, a more accurate classifier ($h_1$, left pane, learned using log-loss) may produce lower {\em team utility} than a less accurate model ($h_2$, right pane).  Suppose the human can either quickly {\em accept} the AI's recommendation or {\em solve} the task themselves, incurring a cost $\lambda$ in time or effort, to yield a more reliable result. The payoff matrix describes the utility of different outcomes. 
    We explore the policy where humans accept recommendations when the AI is confident, but verify uncertain predictions (shown in the light grey region surrounding each hyperplane).  While $h_2$ is less accurate than $h_1$ (because B is incorrectly classified), it results in a higher team utility: Since $h_2$ moved $A$ outside the verify region, there are more {\em correctly classified} inputs on which the user can rely on the system.}
    \label{fig:landing}
\end{figure}  

\checknew{We argue that the most accurate model is not {\em necessarily} the best teammate.}
We show this formally, but the intuition is simple. Considering human-human teams, {\em Is the best-ranked tennis player necessarily the best doubles teammate?} Clearly not---teamwork puts additional demands on participants that extend beyond individual performance on tasks, such as ability to complement and coordinate with one's partner. Similarly, creating high-performing human-AI teams may require training AI systems that exhibit additional {\em human-centered} properties, e.g., facilitating appropriate levels of trust and delegation. Implicitly, this is the motivation behind much work in intelligible AI, including efforts aimed at enhancing the understandability of complex AI inference \cite{horvitz-medinfo86}, interpretability of machine-learned models~\cite{caruana-kdd15,weld-cacm19}, and performing post-hoc explanations of the output of models~\cite{Ribeiro2016,lundberg2017unified}. We move beyond such general motivation and highlight the value of developing methods to model and optimize the collaborative process.
 
 For example, consider the scenario when the system generates advice in which it is uncertain. In practice, users are likely to distrust such recommendations, and rightly so, because a low confidence is often correlated with erroneous predictions~\cite{bansal-arxiv20,hendrycks-arxiv18}.
 In this work, we assume that, when systems have low confidence in their inferences,  users will discard the recommendation and {\em solve} the task themselves, incurring a cost based in the required additional human effort.
As a result, team performance depends on the AI accuracy only in the {\em accept region}, \ie, the region where a user is actually likely to rely on AI. The singular objective of optimizing for AI accuracy (\eg, using \logloss) may hurt team performance when the model has fixed inductive bias. Team performance will benefit from improving AI in the accept regions even if at the cost of performance over the complementary {\em solve regions} (Figure~\ref{fig:landing}).
While there exist other aspects of collaboration that can also be addressed via optimization techniques, such as model interpretability, supporting  complementary skills~\cite{wilder-ijcai20}, or enabling learning among partners, the problem we address in this paper is to account for team-based utility as a basis for collaboration. In sum:
\begin{enumerate}
    \item We highlight an important direction in the field of human-centered AI: When paired with a human overseer, the most accurate ML model may not lead to the highest {\em team performance}. Specifically, we consider settings where, during training the system considers humans' mental model of the AI and how they make use of its recommendations. This setting complements recent advances in {\em learning to defer} where systems are trained when to refuse to share a recommendation to the overseer.
    
    \item For a simple yet ubiquitous form of teamwork, we show that \logloss, the most popular loss function for optimizing AI accuracy, can lead to suboptimal team performance and instead propose directly optimizing for human-AI team's utility. During training, the new objective considers and guides AI performance by considering various human and domain parameters, such as human accuracy, cost of human effort, and cost of mistakes.
    
    \item We present experiments on real-world datasets and models that show improvements in expected team utility achieved by our method. We present qualitative analyses to understand how the re-trained model differs from the most accurate AI, and how the improvements in utility change as a function of domain parameters. We conclude with discussing  optimization issues, loss-metric mismatch, and implications for optimizing team performance for more complex human-AI teams.
\end{enumerate}

\section{Problem Description}
\label{sec:problem}

\begin{figure}[t]
    \centering
     \begin{subfigure}[b]{0.9\linewidth}
        \centering
        \includegraphics[width=\linewidth]{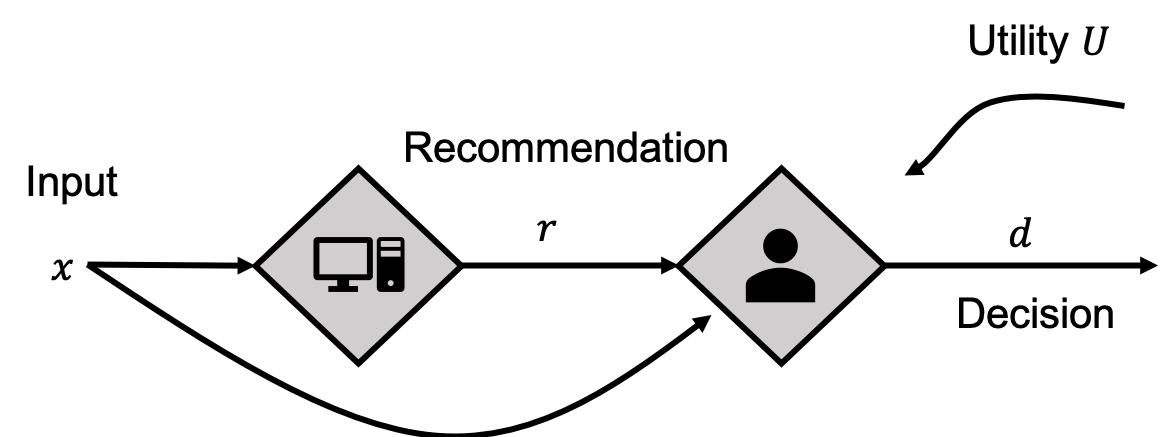}
        \caption{}\label{fig:team}
    \end{subfigure}%
    \quad
     \begin{subfigure}[b]{0.9\linewidth}
        \centering
        \includegraphics[width=\linewidth]{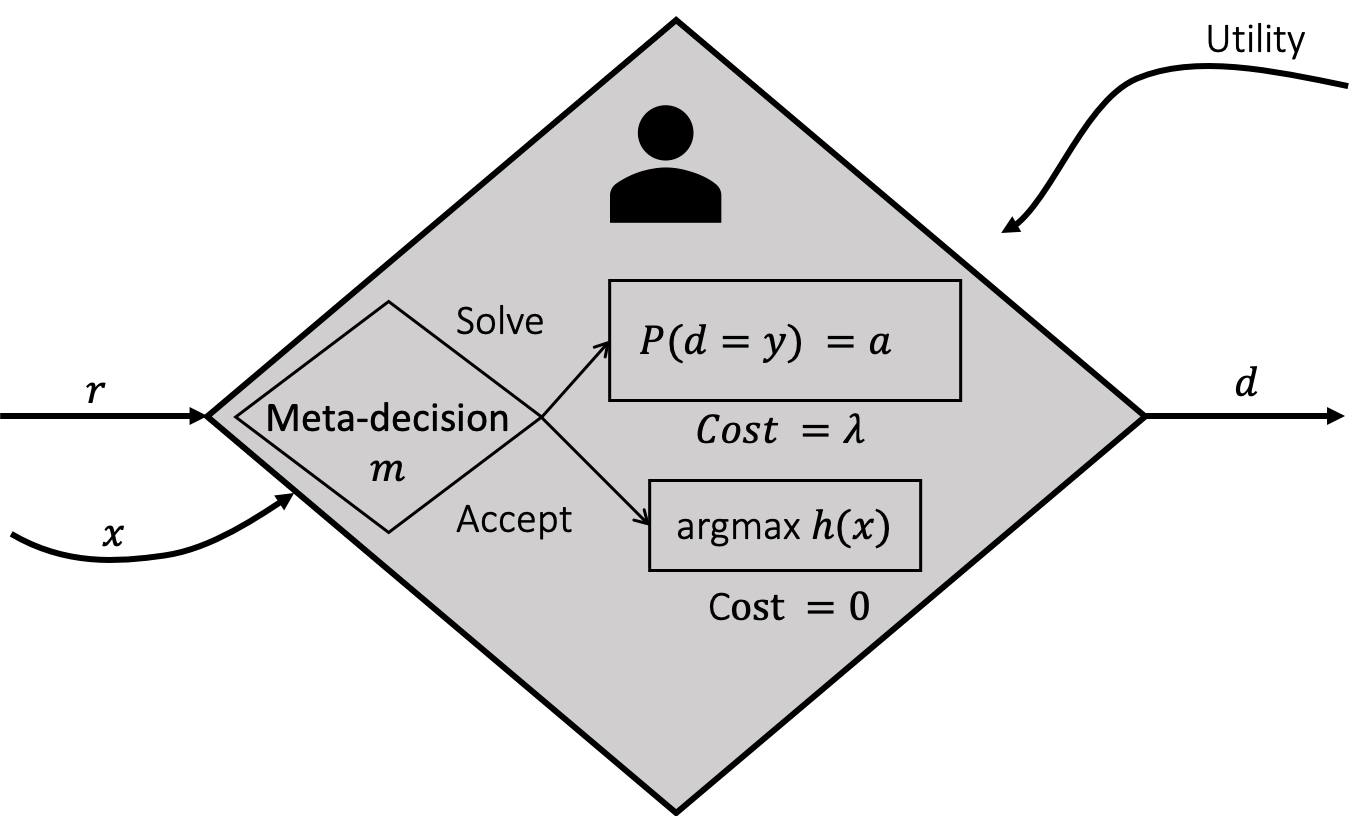}
        \caption{}
        \label{fig:metadecision}
    \end{subfigure}
    
    \caption{(a) AI-advised decision making. (b) To make a decision, the human either accepts or overrides a recommendation. The \override meta-decision is costlier than \accept.}
    \vspace{-0.6em}
\end{figure}

We focus on AI-advised decision making scenarios where a classifier $\classifier$ gives recommendations to a human decision maker to help them make decisions~(Figure~\ref{fig:team}). \checknew{Suppose $\x$ is an $n$-dimensional feature vector (i.e., $X \subset R^n$) and Y is a finite set of possible decisions. For example, for binary classification $Y=\{+, -\}$}. If $\classifier(\x)$ denotes the classifier's output (\ie, a probability distribution over the set of possible outcomes $\labelSpace$) the recommendation $\recommendation$ is a tuple consisting of the predicted label $\predictedLabel = \argmax \classifier(\x)$ and a confidence value $\max \classifier(\x)$, \ie, $\recommendation \coloneqq (\predictedLabel, \max \classifier(\x))$. Using this recommendation, the user computes a final decision $\decisionMaker$. The environment, in response, returns a utility which depends on the quality of the final decision and any cost incurred due to human effort. If the team classifies a sequence of instances, the objective is to maximize the cumulative utility. Before deriving a closed form equation for the objective, we describe the form of the human-AI collaboration we consider along with our assumptions. We study this simple setting as a step to exploring broader opportunities and challenges in team-centric optimization.

\begin{compactenum}
    \item {\em User either accepts the recommendation or solves the task themselves:}  The human computes the final decision by first making a meta-decision $\metaDecisionFunc$: \accept or \override (Figure~\ref{fig:metadecision}). In \accept, the user passes off the AI recommendation as the final decision. In contrast,  in \override, the user ignores the recommendation and computes the final decision themselves. Let $\metaDecisionFunc$ denote the function that maps an input instance and recommendation to a meta-decision in $\metaDecisionSpace=\{\accept, \override\}$.
    Further, $\utilityFunc$ denotes the utility function, which depends on the human meta-decision and final decision $\decisionMaker$ (Figure~\ref{tab:utility}).
    As a result, the optimal classifier $\classifier^*$ would maximize the team's expected utility:
    \begin{equation}
        \classifier^* = \argmax_{\classifier} \expect_{x, y}[\utilityFunc(\metaDecisionFunc, \decisionMaker)]
    \end{equation}
   
    \item {\em Mistakes are costly:} A correct decision results in unit reward. An incorrect decision results in a penalty $\mistakePenalty \geq 1$. 
    
    \item {\em Solving the task is costly:} Since it takes time and effort for the human to perform the task themselves (\eg, cognitive effort), we realistically assume that the \override meta-decision costs more than \accept. Further, without loss of generality, we assume $\overrideCost$ units of cost to \override and zero cost to \accept. Note that even when the cost of \accept is non-zero and the reward for a correct decision is different than one, the utility function can still be transformed and simplified to the same form as in Table~\ref{tab:utility} and be optimized in the same way as we describe henceforth.

    \begin{table}[t]
        \centering
        \footnotesize
        \begin{tabular}{|c|c|c|}
        \hline
          Meta-decision/Decision & Correct & Incorrect \\
             \hline
             \accept [ \acceptS] & $1$ & $-\mistakePenalty$ \\
             \hline
             \override [ \overrideS] & $1 - \overrideCost$ & $-\mistakePenalty-\overrideCost$\\
             \hline
        \end{tabular}
        \caption{Utility as a function of meta-decision and decision.}
        \label{tab:utility}
        \vspace{-1.0em}
    \end{table}
    
    Following the above specifications, we obtain the utility function in Figure~\ref{tab:utility}. The values in the table originate from subtracting the cost of the action from the reward.

    \item {\em Human is uniformly accurate across decisions:} Let $\humanAccuracy \in [0, 1]$ denote the conditional probability that if the user solves the task, they will make the correct decision.
    \begin{equation}
    \label{eq:humanAccuracy}
        P(\decisionMaker = y | \metaDecisionFunc = \overrideS) = \humanAccuracy 
    \end{equation}

    \item {\em Human is rational:} The user chooses the meta-decision that results in highest expected utility. Further, the user trusts the classifier's confidence $\classifier(\x)[\predictedLabel]$ as an accurate indicator of the recommendation's reliability, i.e., true conditional probability of prediction $\predictedLabel$ being correct. As a result, the user will choose \accept if and only if the expected utility of \accept is greater than that of \override.
    \begin{align*}
        \expect[\utilityFunc(\metaDecisionFunc = \acceptS)] &\geq \expect[\utilityFunc(\metaDecisionFunc = \overrideS)]\\
        \classifier(\x)[\predictedLabel] - (1 - \classifier(\x)[\predictedLabel]) \cdot \mistakePenalty &\geq \humanAccuracy - (1 - \humanAccuracy) \cdot \mistakePenalty - \overrideCost\\
       \classifier(\x)[\predictedLabel] &\geq \humanAccuracy - \frac{\overrideCost}{1 + \mistakePenalty}
    \end{align*}
    Let $\threshold$ denote the minimum value of confidence for which the user's meta-decision is \accept.
    \begin{equation}
        \threshold = \humanAccuracy - \frac{\overrideCost}{1 + \mistakePenalty}
    \end{equation}
 This implies the human will follow the following threshold-based policy to make meta-decisions:
\begin{equation}
\label{eq:policy}
    \prob(\metaDecisionFunc = \acceptS) = 
    \begin{cases}
    1 & \text{if}\ \classifier(\x)[\hat{y}] \ge \threshold \\
    0 & \text{otherwise}
    \end{cases}
\end{equation}
\end{compactenum}

\begin{figure}[ht]
 \centering
\includegraphics[width=0.9\linewidth]{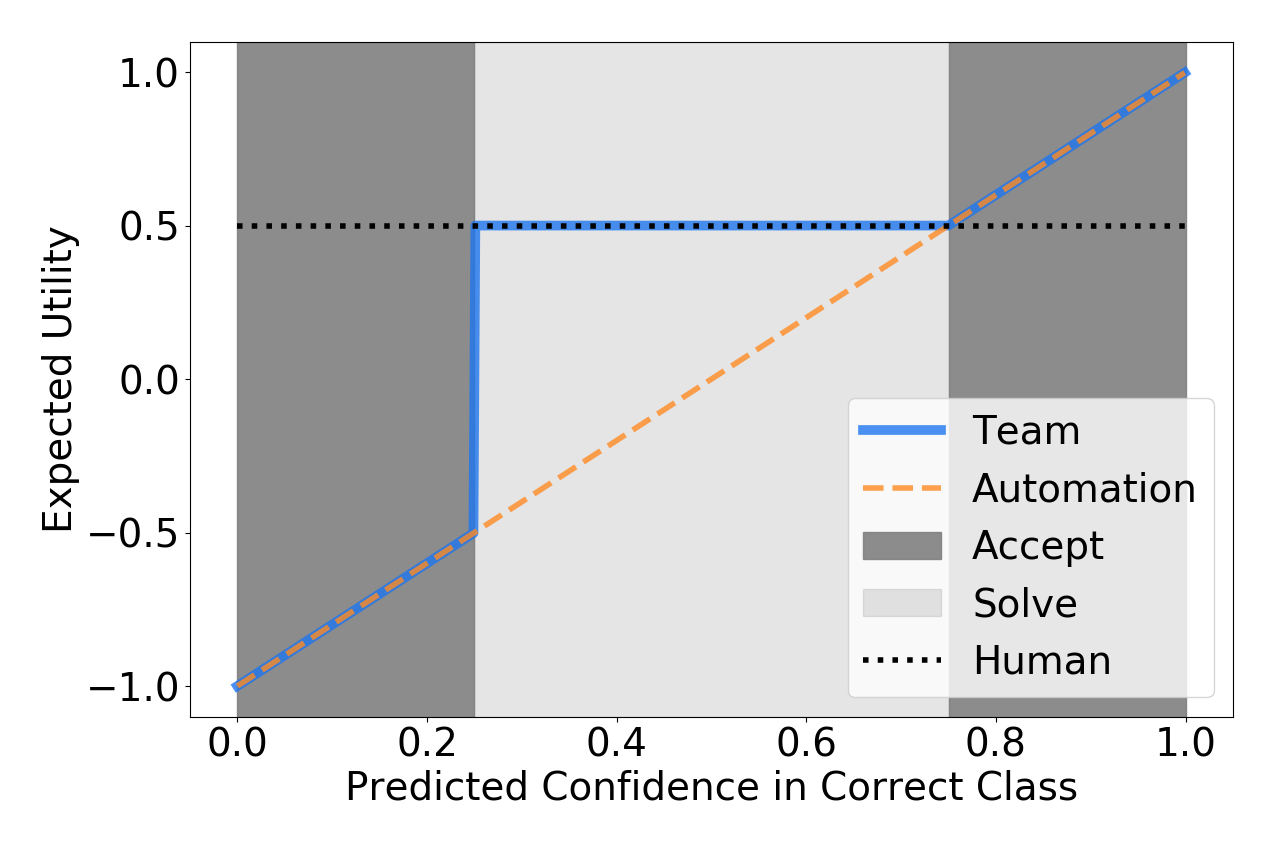}
\caption{Visualization of expected utility when $\overrideCost=0.5, \mistakePenalty=1$, and $\humanAccuracy = 1$ (\ie, the human is perfectly accurate but it costs them half a unit of utility to solve the task). In the \accept region, expected utility of the team is equal to expected utility of the automation, while in the \override region it equals to the human utility. The negative team utility in the left-most region results from over-confident but incorrect recommendations to the human.}
\label{fig:expectedUtility}
\end{figure}

\subsection{Expected Team Utility}

We now derive the equation for expected utility of recommendations for the teamwork that we described above. 
Let $\euFunc$ denote the expected team utility on a given example.

\begin{equation*}
\begin{split}
\euFunc(\x, \correctLabel) &= \expect[\utilityFunc(\metaDecisionFunc, \decisionMaker)]\\
  &= \prob(\metaDecisionFunc=\acceptS)
   \cdot \Big[ \prob(\decisionMaker = y| \metaDecisionFunc=\acceptS) \cdot 1\\
   &~~~~+ \prob(\decisionMaker \neq y| \metaDecisionFunc=\acceptS) \cdot (-\mistakePenalty))\Big]\\
&~~~~+ \prob(\metaDecisionFunc=\overrideS) \cdot \Big[ \prob(\decisionMaker = y| \metaDecisionFunc=\overrideS)\cdot(1 - \overrideCost)\\ 
   &~~~~+ \prob(\decisionMaker \neq y | \metaDecisionFunc=\overrideS) \cdot (-\mistakePenalty - \overrideCost) \Big]
\end{split}
\end{equation*}

Since upon \accept, the human returns the classifier's recommendation, the probability that the final decision is correct is the same as the classifier's predicted probability of the correct decision, \ie, $\prob(\decisionMaker = y| \metaDecisionFunc = \acceptS) = \classifier(\x)[\correctLabel]$.
Substituting this and Equation~\ref{eq:humanAccuracy} we obtain:
\begin{equation*}
\begin{split}
\label{eq:expectedUtility}
    \euFunc(\x, \correctLabel)  &= \prob(\metaDecisionFunc=\acceptS) \cdot \Big[(1 + \mistakePenalty) \cdot \classifier(\x)[\correctLabel] - \mistakePenalty \Big]\\
    &~~~~+ \prob(\metaDecisionFunc=\overrideS) \cdot \Big[(1 + \mistakePenalty) \cdot
     \humanAccuracy - \mistakePenalty - \overrideCost \Big]\\
      &= \prob(\metaDecisionFunc=\acceptS) \cdot \Big[(1 + \mistakePenalty) \cdot (\classifier(\x)[\correctLabel] - \humanAccuracy) + \overrideCost \Big]\\
    &~~~~+ \underbrace{\Big[(1 + \mistakePenalty) \cdot \humanAccuracy - \mistakePenalty - \overrideCost \Big]}_{\text{constant}}
\end{split}
\end{equation*}

Substituting human policy (Equation~\ref{eq:policy}) we obtain:
\begin{equation}
\label{eq:eu}
    \euFunc(\x, \correctLabel) = 
    \begin{cases}
    (1 + \mistakePenalty) \cdot \classifier(\x)[\correctLabel] - \mistakePenalty & \text{if}\ \classifier(\x)[\predictedLabel] \ge \threshold \\
    (1 + \mistakePenalty) \cdot \humanAccuracy - \mistakePenalty - \overrideCost & \text{otherwise}
    \end{cases}
\end{equation}

Figure~\ref{fig:expectedUtility} visualizes the expected team utility of the classifier predictions as a function of confidence in the true label. We convert expected utility into a loss function by negating it, \ie, $- \euFunc(\x, \correctLabel)$.

\section{Experiments}
Experiments in this section address the following questions:
{\bf
\begin{enumerate}
    \item[RQ1] Can we train a classifier with higher utility than the most accurate classifier?
    \item[RQ2] How does the new model qualitatively differ from the most accurate model?
    \item[RQ3] How do the properties of the task affect improvements in utility (\eg, human skill and cost of mistake)?
\end{enumerate}
}

\noindent {\bf Datasets}  We experimented with two synthetic datasets and four real-world binary classification datasets: German credit lending dataset~\cite{data-hofman}, FICO credit risk assessment~\cite{data-fico}, recidivism prediction~\cite{data-recidivism}, and MIMIC-3 mortality prediction~\cite{harutyunyan-sd19}. The real datasets are drawn from high-stakes domains where machine learning has already been deployed or has been discussed being employed to assist human decision makers. On  the synthetic datasets, \toy dataset refers to a dataset we created by sampling 10000 points from the data distribution similar to Figure~\ref{fig:landing}. Moons refers to the classic two moons non-linear classification problem.\footnote{\url{https://scikit-learn.org/stable/modules/generated/sklearn.datasets.make_moons.html}}

\begin{table}[t]
\footnotesize
    \centering
    \begin{tabular}{|l|r|r| r|}
    \hline
        {\bf Dataset} & {\bf \#Features} & {\bf Size} & {\bf Frac. Pos.} \\
        \hline
        \toy & 2 & 10000 & 0.43 \\
        Moons & 2 & 10000 & 0.50\\
        German & 24 & 1000 & 0.30\\
        Fico & 39 & 9861 & 0.52\\
        Recidivism & 13 & 6172 & 0.46\\
        MIMIC & 714 & 21139 & 0.13\\
        \hline
    \end{tabular}
    \caption{Number of features and size of binary classification datasets used for experiments. The original Fico dataset contains 23 features but 39 after preprocessing categorical features into binary features.}
    \label{tab:datasets}
\end{table}

\noindent {\bf Model Training}
We experimented with two types of models: logistic regression and multi-layered perceptron (two hidden layers with 50 and 10 units).
For each task (defined by a choice of task parameters, dataset, model, and loss) we optimized the loss using the Adam optimizer and also used standard, well-known training practices such as  regularization, check-pointing the model best validation performance, and learning rate schedulers. We selected the best hyperparameters using five-fold cross validation, including values for the learning rate, batch size, patience, decay factor of the learning rate scheduler, and the L2 regularization weight. (Range of parameters detailed in the Appendix).

In initial experiments to optimize team utility, we observed that the classifier's loss (in this case, negative of expected utility) remained constant over the optimization process. This happened because, in practice, random initializations resulted in classifiers that were uncertain on most of the data distributions considered. By definition, the expected utility is flat and constant in regions of uncertainty (see Figure~\ref{fig:expectedUtility}). Thus, the gradient was zero and uninformative over these ranges. To overcome this issue, we initialized the classifiers with the (already converged) most accurate classifier.

\begin{table*}[ht]
\footnotesize
\centering

\begin{tabular}{|ll|l|l|l|l|l|l|}
\hline
 &  & \multicolumn{3}{c|}{\textbf{Logloss}} & \multicolumn{3}{c|}{\textbf{Expected Utility Loss}} \\ \hline
\multicolumn{1}{|l|}{\textbf{Classifier}} & \textbf{Dataset} & \textbf{Accuracy} & \textbf{Expected Util.} & \textbf{Emp. Util.} & \textbf{$\Delta$ Accuracy} & \textbf{$\Delta$ Expected Util.} & \textbf{$\Delta$ Emp. Util.} \\ \hline
\multicolumn{1}{|l|}{\multirow{6}{*}{Linear}} & Fico & 0.729 & 0.487 & 0.575 & -0.247 & \textbf{0.013} & -0.075 \\
\multicolumn{1}{|l|}{} & German & 0.754 & 0.529 & 0.594 & -0.015 & 0 & -0.019 \\
\multicolumn{1}{|l|}{} & MIMIC & 0.881 & 0.694 & 0.8 & -0.004 & \textbf{0.066} & -0.035 \\
\multicolumn{1}{|l|}{} & Moons & 0.885 & 0.687 & 0.79 & -0.02 & \textbf{0.079} & -0.006 \\
\multicolumn{1}{|l|}{} & recidivism & 0.669 & 0.485 & 0.52 & -0.17 & \textbf{0.015} & -0.02 \\
\multicolumn{1}{|l|}{} & Scenario1 & 0.858 & 0.524 & 0.593 & -0.165 & \textbf{0.102} & \textbf{0.061} \\ \hline
\multicolumn{1}{|l|}{\multirow{6}{*}{MLP}} & Fico & 0.725 & 0.472 & 0.574 & -0.244 & \textbf{0.028} & -0.074 \\
\multicolumn{1}{|l|}{} & German & 0.752 & 0.53 & 0.618 & -0.036 & -0.027 & -0.056 \\
\multicolumn{1}{|l|}{} & MIMIC & 0.881 & 0.719 & 0.799 & -0.001 & \textbf{0.049} & -0.029 \\
\multicolumn{1}{|l|}{} & Moons & 1 & 0.944 & 0.989 & 0 & \textbf{0.049} & \textbf{0.006} \\
\multicolumn{1}{|l|}{} & Recidivism & 0.674 & 0.467 & 0.521 & -0.168 & \textbf{0.033} & -0.021 \\
\multicolumn{1}{|l|}{} & Scenario1 & 1 & 0.826 & 0.854 & -0.1 & \textbf{0.08} & \textbf{0.057} \\ \hline
\end{tabular}
    \caption{Comparison of accuracy, expected and empirical team utilities of classifiers optimized for \logloss (with a checkpoint on accuracy) and expected team utility (with a checkpoint on expected utility) using Adam for $\overrideCost=0.5$, $\humanAccuracy=1.0$, $\mistakePenalty=1.0$. Observations averaged over 50 train/test splits. $\Delta$ indicates difference with respect to \logloss. Classifier trained to optimize expected team utility achieves higher expected utility at the cost of automation accuracy. However, we notice a mismatch between expected and empirical utilities-- empirical utility {\em decreased} even though expected utility increased.}
    \label{tab:performance}
\end{table*}

\noindent {\bf Metrics: Empirical and Expected Utility} We evaluated our systems on two metrics of team utility: expected team utility (Equation~\ref{eq:eu}) and empirical team utility, which draws discrete rewards from the pay-off described in Table~\ref{tab:utility}. 
A key difference between expected and empirical utilities is that the former incentivizes systems that output a calibrated belief, \ie, in the \accept region it assigns a score proportional to the system's confidence in the correct class (Figure~\ref{fig:expectedUtility}). Empirical utility, in contrast, does not differentiate between a low- and a high-confidence recommendation in the \accept region as long as they are both correct (or both are incorrect).

Each metric offers different advantages. Maximizing empirical utility aligns well with existing non-probabilistic discrete metrics for evaluating ML classifiers (such as, accuracy, F1-score, and AUPRC), which exclusively focus on the discriminative power of models. In contrast, maximizing expected utility is critical for decision making under uncertainty, \ie, when the outcome of decisions may be probabilistic and thus a rational agent should maximize for its decision's expected utility.  In fact, the primary result of utility theory, the accepted, normative theory of action under uncertainty, is that ideal decisions are those that maximize expected utility~\cite{morgenstern-theory}. Maximizing expected utility requires the use of calibrated probabilities, which is an aspect that is not reflected in empirical utility. Moreover, expected utility optimization is useful in cases when empirical evaluation of metrics is not feasible due to delayed reward in the real world or when the definition of empirical ground truth labels is soft and non-discrete.

\subsection{Results}

\begin{table*}[t]
\footnotesize
\centering
\begin{tabular}{|l|l|l|l|l|l|}
\hline
{\bf Dataset}          & \bf{ Expected Util {\tiny LL}} & {\bf Emp. Util {\tiny LL}} & {\bf $\Delta$Expected Util (A)} & {\bf $\Delta$ Emp. Util (B)} & {\bf $\Delta^*$ Emp. Util (C)} \\
\hline
Fico-2d       & 0.475            & 0.511          & 0.025                      & -0.011                   & -0.004                   \\
German-2d     & 0.514            & 0.6            & 0.076                      & -0.004                   & -0.016                   \\
MIMIC-2d      & 0.641            & 0.772          & 0.121                      & -0.009                   & 0.005                    \\
Moons             & 0.767            & 0.813          & 0.016                      & -0.006                   & 0.034                    \\
Recidivism-2d & 0.478            & 0.518          & 0.022                      & -0.017                   & 0.007                    \\
Scenario1         & 0.707            & 0.715          & 0.045                      & 0.069                    & 0.068                   \\
\hline
\end{tabular}

\caption{Test performance of linear classifier that optimizes \logloss and team utility using brute-force optimization on two-dimensional domains. While we observe consistent improvements in the team's expected utility (column marked A) across domains, improvements in expected utility did not translate to improvements in {\em empirical utility} (values in column marked B are negative), indicating a mismatch between the expected and empirical metrics of team utilities. At the same time, exhaustive search shows existence of linear classifiers with higher empirical utility (column marked C). Values were averaged over five seeds. Observations in column C on Fico-2d and German-2d were negative on test set due to over-fitting.}
\label{tab:exhaustive}
\end{table*}

\noindent {\bf RQ1:} Table~\ref{tab:performance} shows that the new classifier can improve expected team utility over \logloss. These improvements are often achieved by sacrificing the classifier's individual accuracy. For example, on \toy the new linear classifier improved expected utility from 0.524 to 0.606 even though it was less accurate.

When we considered {\em empirical} utility, our method did not always result in improvements. For example, for the linear classifier, while on \toy, the empirical utility increased from 0.593 to 0.654, but on MIMIC it decreased from 0.8 to 0.765.
Ideally, one would expect that an increase in expected team utility would be accompanied with proportional increase in empirical team utility.
However, as Table~\ref{tab:performance} shows, this was often not the case.

While this mismatch between empirical and expected utilities seems counterintuitive, it is a well know problem; \citet{huang2019addressing} noticed a mismatch between various common ML evaluation metrics, such as \logloss, zero-one loss, and AUPRC.
However, we still considered the possibility that, in practice, the mismatch perhaps resulted from stochastic optimization getting stuck in local minimas, and that a better optimization procedure would alleviate this mismatch. 
To pursue this conjecture, we developed two-dimensional versions of our dataset (by selecting two top most informative features) and trained linear classifiers using exhaustive search, which by definition cannot get stuck in local minimas. We again found a persistence of the mismatch between expected and empirical utilities (Table~\ref{tab:exhaustive}). In addition, we also noticed that there exist classifiers with higher empirical utility if the exhaustive search maximizes directly for empirical utility (column C in Table~\ref{tab:exhaustive}), which further demonstrates the existence of the mismatch. \footnote{Note that directly optimizing for empirical utility is not effective via stochastic optimization.}

These results provide evidence that the challenge with achieving comparable increases in empirical utility to those in expected utility is not only due to optimization issues (\eg, local minimas and plateaus due to flatness of the expected utility curve in the \override region). There exists a fundamental ML challenge of loss-metric mismatch, which was prominent in our setup. In the rest of the section, we present further analyses of improvements in the normative decision making metric of expected utility, which as described earlier, is useful in decision-making under uncertainty.

\noindent {\bf RQ2:} While the metrics in Table~\ref{tab:performance} (change in accuracy and utility) provide a global understanding of the classifier behavior, here we attempt to understand {\em how} these improvements were achieved and whether the behavior of the new models is consistent with the original intuition. Figure~\ref{fig:behaviorToy} displays the difference in behavior (averaged over 50 seeds) between the classifiers produced by \logloss and the one that maximizes team utility on the \toy and MIMIC dataset. Specifically, as shown in Figure~\ref{fig:behaviorToy}, we visualize and compare the following behaviors of the two classifiers:

\begin{figure}[t]
    \centering
    \includegraphics[width=\columnwidth]{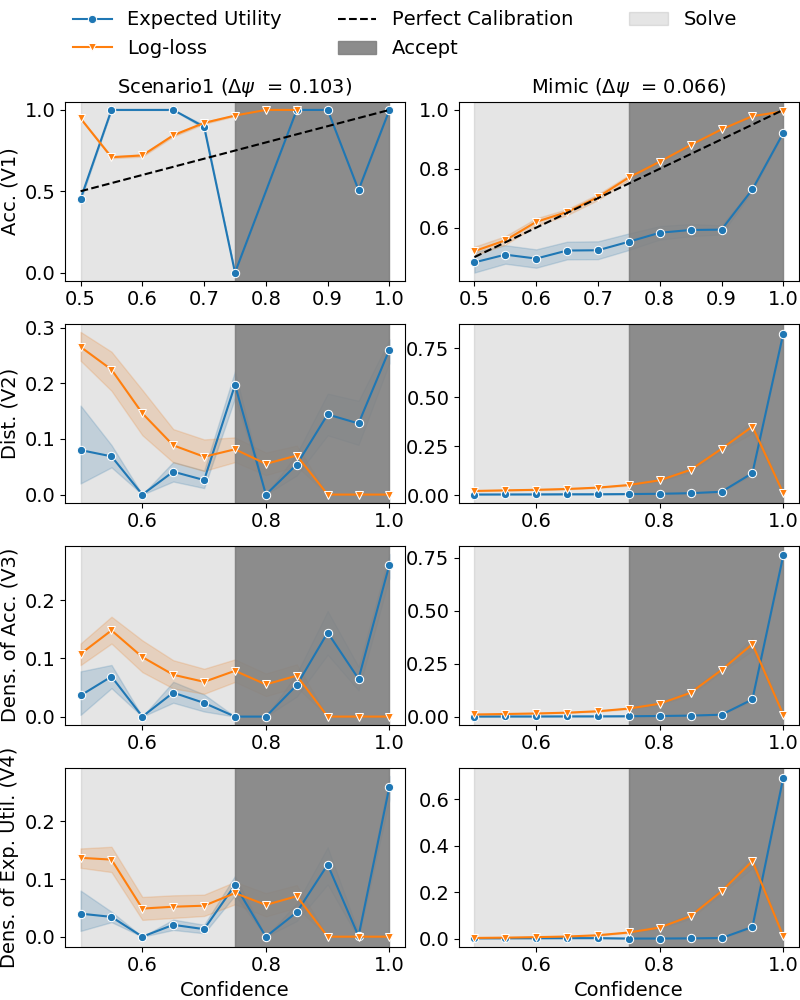}
    \caption{Behavior of linear classifiers that optimize \logloss and expected team utility on the \toy and MIMIC datasets (observations averaged over 50 runs). 
    The latter makes fewer predictions in the \override region and also sacrifices accuracy in that region to increase it in \accept. We observed a behavior similar for the MLP model on all datasets (ommitted due to space constraints).} 
    \label{fig:behaviorToy}
\end{figure}

\begin{itemize}
    \item[V1.] Calibration using {\em reliability curves}, which compare system confidence and its true accuracy. A perfectly calibrated system, for example, will be 80\% accurate on regions that is 80\% confident. However, in practice, systems may be over- or under-confident.
    \item[V2.] Distributions of confidence in predictions. For example, in Figure~\ref{fig:behaviorToy}, the new classifier makes more high-confidence predictions than the most accurate classifier.
    \item[V3.] Density of system accuracy as function of confidence in true label. Thus, the area under this curve indicates the system's total accuracy. Note that, for our setup, the area under the curve in the \accept region is more crucial.
    \item[V4.] Density of expected utility as a function of confidence.
\end{itemize}

The classifier optimized for the team's expected utility results in dramatically different predictions than the classifier trained using \logloss: The new classifier sacrifices accuracy on the uncertain examples (\override region) to make higher numbers of high-confidence predictions (\accept region). Most importantly, it also increases the density of system accuracy in the \accept region, which is where the system accuracy matters and contributes to team utility. Figure~\ref{fig:behaviorToy} illustrates the same behavior on the MIMIC in-hospital mortality prediction dataset.

An interesting exception was Fico, where the system learned to always be uncertain.
This may make sense for the Fico domain because, as shown in Table~\ref{tab:performance}, even though the most accurate linear classifier is 73\% accurate on Fico, it achieves an expected team utility of 0.487. This is less than the expected utility achieved if humans solved the task alone. Hence, the more accurate classifier leads to lower expected team utility.  We observed a similar behavior on recidivism prediction where the linear classifier led to team performance lower than that associated with  people making decisions unaided, even though the classifer had a 67.4\% accuracy (Table~\ref{tab:performance}). These cases illustrate timely concerns and questions of when and if an AI should be deployed to assist human decision-making, which we further discuss in the ethical statement.

\begin{table}[t]
\footnotesize
\centering
\begin{tabular}{|l|l|l|l|}
\hline
\textbf{dataset} & \textbf{a=0.8} & \textbf{a=0.9} & \textbf{a=1} \\ \hline
Fico & 0.257 (0.133) & 0.337 (0.071) & 0.487 (0.013) \\
German & 0.397 (0.046) & 0.444 (0.035) & 0.529 (0) \\
MIMIC & 0.625 (0.127) & 0.644 (0.111) & 0.694 (0.066) \\
Moons & 0.582 (0.162) & 0.616 (0.139) & 0.687 (0.079) \\
Recidivism & 0.155 (0.073) & 0.292 (0) & 0.485 (0.015) \\
Scenario1 & 0.224 (0.324) & 0.364 (0.248) & 0.524 (0.102) \\ \hline
\end{tabular}
\caption{Expected utility of \logloss and improvements for linear classifiers ($\Delta$ Expected Util. shown in brackets) with varying human accuracy ($\humanAccuracy$) and ($\overrideCost=0.5$ and $\mistakePenalty=1.0$). Results averaged over 50 random seeds. Improvements in expected utility are higher when the human is less accurate.}
\label{tab:varyAccuracy}
\end{table}

\begin{table}[t]
\footnotesize
\centering
\begin{tabular}{|l|l|l|l|}
\hline
\textbf{dataset} & \textbf{$\mistakePenalty$=1} & \textbf{$\mistakePenalty$=3} & \textbf{$\mistakePenalty$=5} \\ \hline
Fico & 0.487 (0.013) & 0.474 (0.026) & 0.481 (0.019) \\
German & 0.529 (0) & 0.427 (0.057) & 0.367 (0.118) \\
MIMIC & 0.694 (0.066) & 0.58 (0.008) & 0.543 (0) \\
Moons & 0.687 (0.079) & 0.637 (0.065) & 0.594 (0.085) \\
Recidivism & 0.485 (0.015) & 0.495 (0.004) & 0.498 (0.001) \\
Scenario1 & 0.524 (0.102) & 0.501 (0.02) & 0.5 (0) \\ \hline
\end{tabular}
\caption{Expected uility of \logloss and improvements for linear classifiers (\ie, $\Delta$ Expected Util., shown in brackets) with varying cost of mistakes ($\mistakePenalty$) and ($\overrideCost=0.5, \humanAccuracy=1.0$). Results averaged over 50 random seeds. On most datasets, gains diminish as the cost of mistakes increases.}
\label{tab:varyPenalty}
\end{table}

\noindent {\bf RQ3:} Since properties such as the accuracy of users and penalty of mistakes may be task-dependant (\eg, an incorrect diagnosis may be costlier than incorrect loan approval), we varied human accuracy $\humanAccuracy$ and mistake penalty $\mistakePenalty$ to study the sensitivity in improvements in team utility to a wider range of these task parameters.

Table~\ref{tab:varyAccuracy} shows improvements in expected utility as we vary human accuracy from 80\% to 100\% while keeping $\overrideCost$ and $\mistakePenalty$ constant to 0.5 and 1, respectively. These three values of $\humanAccuracy$ result in three new values of optimal threshold $\threshold$: 0.55, 0.65, and 0.75, thus gradually expanding the confidence region in which the user is likely to \override because they themselves are more accurate. We notice higher improvements in expected utility from deploying a system when humans are less accurate, \eg, Table~\ref{tab:varyAccuracy} shows that, on Fico, improvement in expected utility is 0.133 when the human is 80\% accurate whereas it is 0.013 when they are perfect. One explanation for this behavior is that when humans are less accurate there is greater value from system recommendations, which widens the \accept region and increases the scope where the AI can provide value to the team.

Similarly, Table~\ref{tab:varyPenalty} shows the impact of varying cost of mistakes $\mistakePenalty$ on improvements. The three values of $\mistakePenalty$ increase the \accept threshold gradually from 0.75 to 0.91, and therefore shrink the size of the \accept region. Hence, we start observing smaller gains when the cost of mistake is high, \eg, on the MIMIC dataset there are no gains, although the trend is also subject to the shape of expected utility and how easy it is to optimize it. In overall, the trend emphasizes once again that for extremely high-stake decisions, automation or AI recommendation may not always provide value.

\section{Discussion and Future Work}
\checknew{
\noindent {\bf Implications for complex human-AI teams

}
While we investigated a simplified human-AI teamwork (as defined in Section~\ref{sec:problem}), our
setup allows extensions to more complex team and users. For example, one can relax our assumption that users are rational  by modifying the human-policy in Equation~\ref{eq:policy}, so that when the prediction confidence is greater than the threshold, the user chooses \accept with probability $p<1$, instead of 1.0. Here, $1-p$ denotes the probability of the user being irrational --- assessed from historical data, if available. 
Similarly, in more complex situations users may make \accept and \override decisions using a richer, more complex mental model instead of relying on just model confidence.}
Such scenarios are common in cases where the system confidence is an unreliable indicator of performance (\eg, due to poor calibration), and, as a result, the user develops an understanding of system failures in terms of domain features. For example, Tesla drivers may learn to override the Autopilot considering such features as road, sun glare, and weather conditions. We can reduce the case where users have a complex mental model to the policy that we studied. 
\checknew{Specifically, we can construct a new loss function in terms of human utility (in this case, constant) when the prediction belongs to the \override region (as described by the user's mental model) and automation utility otherwise. 

While the above extensions to our model are a start, even they may present challenges-- If we cannot optimize empirical utility for our simplified case, it may be harder to optimize performance in the extensions as they contain more complex user behavior and the resultant loss surface is likely to be more complex, containing combinations of plateaus and local optima. In addition to these extensions, future work should also consider more general uses of AI recommendations in support of human decision making.  For example, we need to consider common uses that are not constrained to policies where a user either accepts an AI recommendation or relies completely on their own reasoning. It is natural to expect that users in human-AI teams will employ their own {\em evidential reasoning} to fuse AI inferences (and associated confidences if shared) with their own assessments. Furthermore, user's mental models may not be static; instead, they may change with time as users learn more about the AI. Mental models may also vary across users, as different people might have different propensity to accept machine recommendations.

}

\checknew{
Human-subject evaluations are an important next step to understand how factors such as biases, variations in user expertise, irrational behavior come to play in practice. Is our simple model of human behavior sufficient for our approach to yield gains in practice?  We view our work as a fundamental first step showing the potential impact of a human-centered model and motivating additional work including real-world studies with human subjects.
Over time, we hope to learn and incorporate rich  (and individualized) models of human behavior into our framework and test them in real-world human-AI teams. 
}

\paragraph{Empirical utility and auxiliary loss functions}
While optimizing for teamwork, we faced two fundamental optimization challenges. First, we observed an an inherent mismatch between empirical and expected utility as shown in the exhaustive experiments for two dimensional data, which hindered optimization on the empirical metric, which is often a central consideration in ML. 
Second, current optimization techniques were not always effective and in fact sometimes they did not change model behavior because the optimization approaches got stuck due to zero gradients and local minimas \override region.

\begin{figure}
    \centering
    \includegraphics[width=\linewidth]{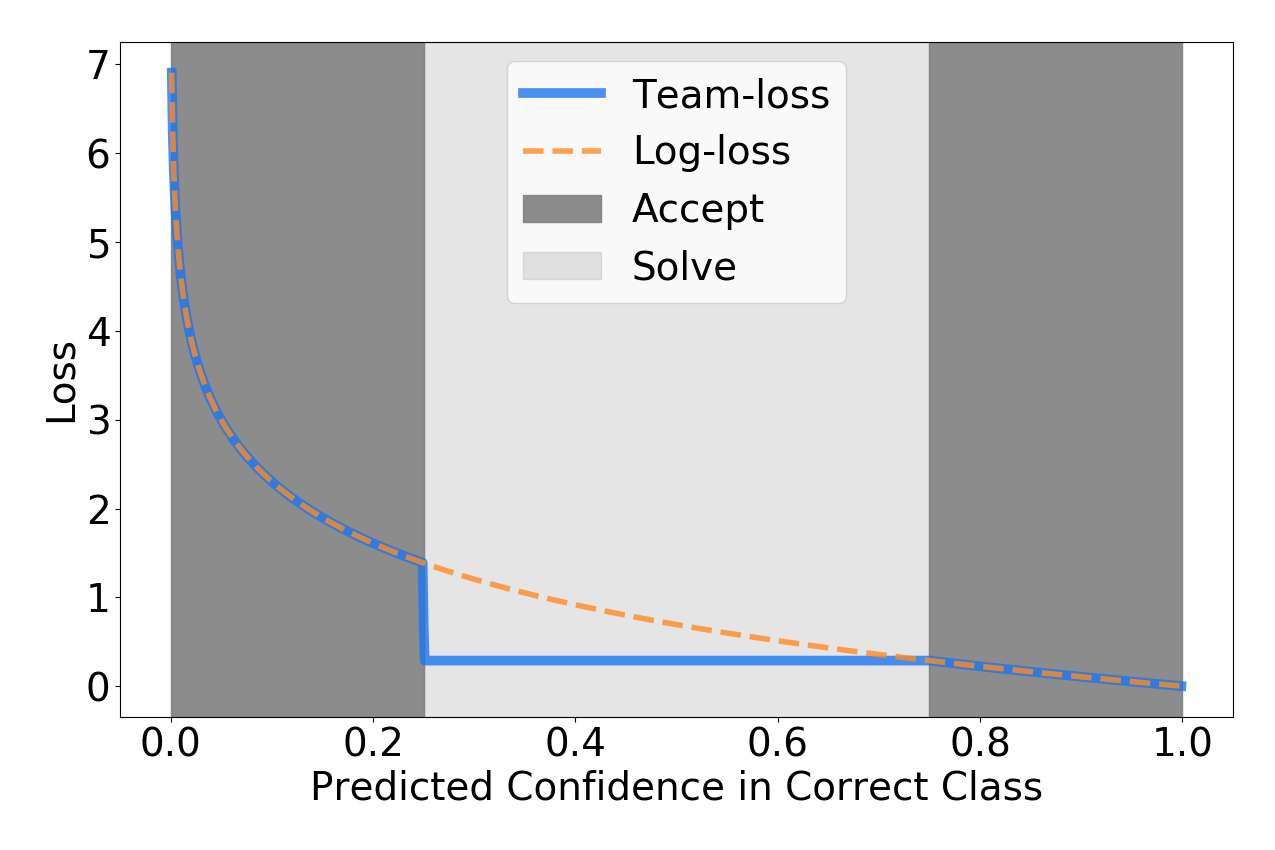}
    \caption{An example of an auxiliary loss function Team-loss defined as $=log(\euFunc(\x, \correctLabel) + K)$, which is equal to Log-loss in \accept region and constant otherwise. Here, $K$ is a positive constant we added so that the logarithmic is valid.}
    \label{fig:nelu}
\end{figure}

To support empirical utility maximization, in our initial analysis, we also experimented with an auxiliary loss function, as shown in Figure~\ref{fig:nelu}. However, in our experiments this loss function did not lead always lead to significant gains in empirical utility and when it did it only lead to marginal improvements. Based on these theoretical and practical challenges, we invite future work on machine learning optimization and human-AI collaboration to develop new optimization techniques that work well beyond \logloss, robustly over a more general set of loss of functions that can capture team utility.

\checknew{
\paragraph{Mental models and explainable AI}

To increase team performance we focused on adapting one AI property to  user mental models--  the AI should be more accurate on instances when the user is more likely to trust model recommendations. Similarly, future work should study whether other aspects of human-AI collaboration can be improved by considering user mental models. For instance, mental models could help inform the nature of explanation given to the users. Such as, when users already trust the model, the system may be better off offering concise explanations. In contrast, when the users are likely to distrust the model, they system should be ready to offer detailed explanations/arguments supporting its prediction~\cite{bansal-arxiv20}.
Eventually, work on explainable AI aims to improves human-AI collaboration by providing a layer of communication between users and AI systems. Since explainability does not guarantee improvements in collaboration~\cite{bansal-arxiv20}, there is need to bring collaboration as an objective to every step of system development, starting from the training objective. We hope that this work paves the work for future directions towards uncovering how to develop AI systems for collaboration.
}
\section{Related Work}

Our approach is closely related to {\em maximum-margin classifiers}, such as an SVM optimized with the hinge loss~\cite{burges1998tutorial}, where a larger soft margin can be used to make high-confidence and accurate predictions. However, unlike our approach, it is not possible to directly plug the domain's payoff matrix (\eg, in  Table~\ref{tab:utility}) into such a model. Furthermore,  the SVM's output and margin do not have an immediate probabilistic interpretation, which is crucial for our problem setting. One possible (though computationally intensive) solution direction is to convert margin into probabilities, \eg, using post-hoc calibration (\eg,  Platt scaling~\cite{platt-99}), and use cross-validation for selecting margin parameters to optimize team utility. While it is still an open question whether such an approach would be effective for SVM classifiers, in this work we focused our attention on gradient-based optimization.

Another related problem is {\em cost-sensitive learning}, where different mistakes incur different penalties; for example, false-negatives may be costlier than false-positives~\cite{zadrozny2003cost,bach-jmlr06}. A common solution here is up-weighting the inputs where the mistakes are costlier.  Also relevant is work on {\em importance-based learning} where re-weighting helps learn from imbalanced data or speed-up training.  However, in our setup, re-weighting the inputs makes less sense--- the weights would depend on the classifier's output, which has not been trained yet. An iterative approach may be possible, but our initial analysis showed this approach is prone to oscillations. We leave exploring this avenue for future work.

A fundamental line of work that renders AI predictions more actionable (for humans) and better suitable for teaming is {\em confidence-calibration}, for example, using Bayesian models~\cite{ghahramani2015probabilistic,beach1975expert,gal_dropout_2016} or via {\em post-hoc} calibration~\cite{platt-99,zadrozny2001obtaining,guo2017calibration,niculescu2005predicting}. A key difference between these methods and our approach is that \abbrvlossname\xspace {\em re-trains} the model to improve on inputs on which users are more likely to rely on the AI predictions. The same contrast distinguishes our approach from {\em outlier detection} techniques~\cite{hendrycks2018deep,lee2017training,hodge2004survey}. 

Closely related is research on {\em learning to defer}~\cite{madras-neurips18,mozannar-arxiv20} and {\em learning to complement}~\cite{wilder-ijcai20}, where the classifier can abstain and  defer/query the task to the user, while accounting for  costs and benefits of intervention. While the "Solve" meta-decision in our framework corresponds to the defer action, our work differs from these works in two important ways. First, the defer action in prior work is system-initiated whereas in our case it is user-initiated and based on their mental model. Second, learning to defer does not preclude our methods, since users may create mental models even when the system does not defer and so the team may still benefit from training a model that accounts for user's mental model.

Other recent work that adjusts model behavior to accommodate collaboration includes {\em backward-compatibility for AI}~\cite{bansal2019updates}, where the model considers user interactions with a previous version of the system to preserve trust across updates. Recent user studies showed that when users develop mental models of AI system, properties besides accuracy are also desirable, such as {\em parsimonious} and {\em deterministic} error boundaries \cite{bansal2019beyond}. Our approach is a first step towards implementing these desiderata within ML optimization itself. Other approaches regularize or constrain model optimization for other human-centered requirements such as local- or global-interpretability~\cite{wu2019regional} or fairness~\cite{jung2019eliciting,zafar2015fairness}.

\section{Conclusions}
We studied opportunity to train classifiers that optimize human-AI team performance. We showed the value of optimizing the expected utility of decision making of human-AI teams in contrast to traditional model optimization focusing solely on automation accuracy. Investigations and visualizations of classifier behavior before and after proposed optimization show that the methods can be harnessed to fundamentally change model behavior and improve the team utility. Changes in model behavior include (i) sacrificing model accuracy in low confidence regions for more accurate high-confidence predictions and (ii) increasing accuracy and number of high-confidence predictions. Such behaviors were observed in both synthetic and real-world datasets where AI is known to be employed as support for human decision makers, and across various domain parameters such as human accuracy and cost of mistake.

\clearpage
\subsection{Ethical Statement}
\checknew{
A broader contribution of this work is to rethink how  ML models are defined and optimized when they are deployed in human-AI collaboration scenarios, \eg, for supporting human decision making
in high-stakes areas (including healthcare and criminal justice) where AI systems already influence user decisions with important consequences for individuals and society. Since most AI systems are optimized automation performance, more research is needed to create effective advisory systems by integrating team-centered considerations in the formal machinery of optimization used to build and execute these AI systems. We examined one approach to raising the expected value of AI-aided human decision making by considering teamwork in the optimization objective.

Beyond the direct use of the methods for optimizing human-AI teamwork, the methods can be valuable for building insights on teaming. For example, results showed that there exist regions in the space of collaboration parameters where automated assistance, or even providing an AI recommender, may not make sense-- when the cost of mistakes was high and human decisions are sufficiently accurate, the algorithm always hands over control to humans, discouraging the need for algorithmic support. Such an analysis highlights the importance of carefully questioning and evaluating whether AI deployment is beneficial from a team perspective. A more rigorous evaluation requires robust and online estimation of costs and user behavior to ensure that the training and real-world objectives align. While we did not address the problem of estimating and updating such parameters, we wish to bring attention to the fact that problems such as underestimation of costs (or overestimation of rewards) may still lead to high-cost mistakes even when following the optimization approach we proposed in this paper. We hope that advances in interdisciplinary research on measuring the impact and costs in socio-technical systems will further inform decisions and designs about the role and behavior of AI in human-AI teamwork in future work.  

Finally, we recognize that significant ethical issues are raised by the nature of human oversight and agency over AI in our simplified human-AI teaming. Our formulation of user policy assumed that, when the AI system is confident, the user completely trusts AI inferences and forgoes further human deliberation. Such a policy can lead to inappropriate transfers of responsibility in realistic settings. Even if the model is confident and has been historically correct, humans will still need to stay cognizant of the potential for poorly characterized and unexpected modes of AI failure, \eg, due to distributional shifts  or  changes in the influences of latent variables with changes in context or workload.

Thus, in real-world settings, the policy that we studied can be dangerous. 
In uses of AI, where high-confidence recommendations are typically trusted and there is a practice of little or no human deliberation about the validity of automated output, human overseers of AI should be aware of their reliance and their need to take full accountability for outcomes linked to the inferences.}
\section{Acknowledgments}
This material is based upon research initially performed during Gagan Bansal's summer internship at Microsoft Research, with continuing support by ONR grant N00014-18-1-2193, NSF RAPID grant 2040196, the University of Washington WRF/Cable Professorship, and the Allen Institute for Artificial Intelligence (AI2). The authors thank \checknew{Zeyuan Allen-Zhu, Rich Caruana, Bryan Wilder, and the anonymous reviewers for helpful comments.}
\bibliography{ijcai20}

\newpage
\section{Appendix}

\begin{figure*}[t]
    \centering
    \includegraphics[width=0.8\textwidth]{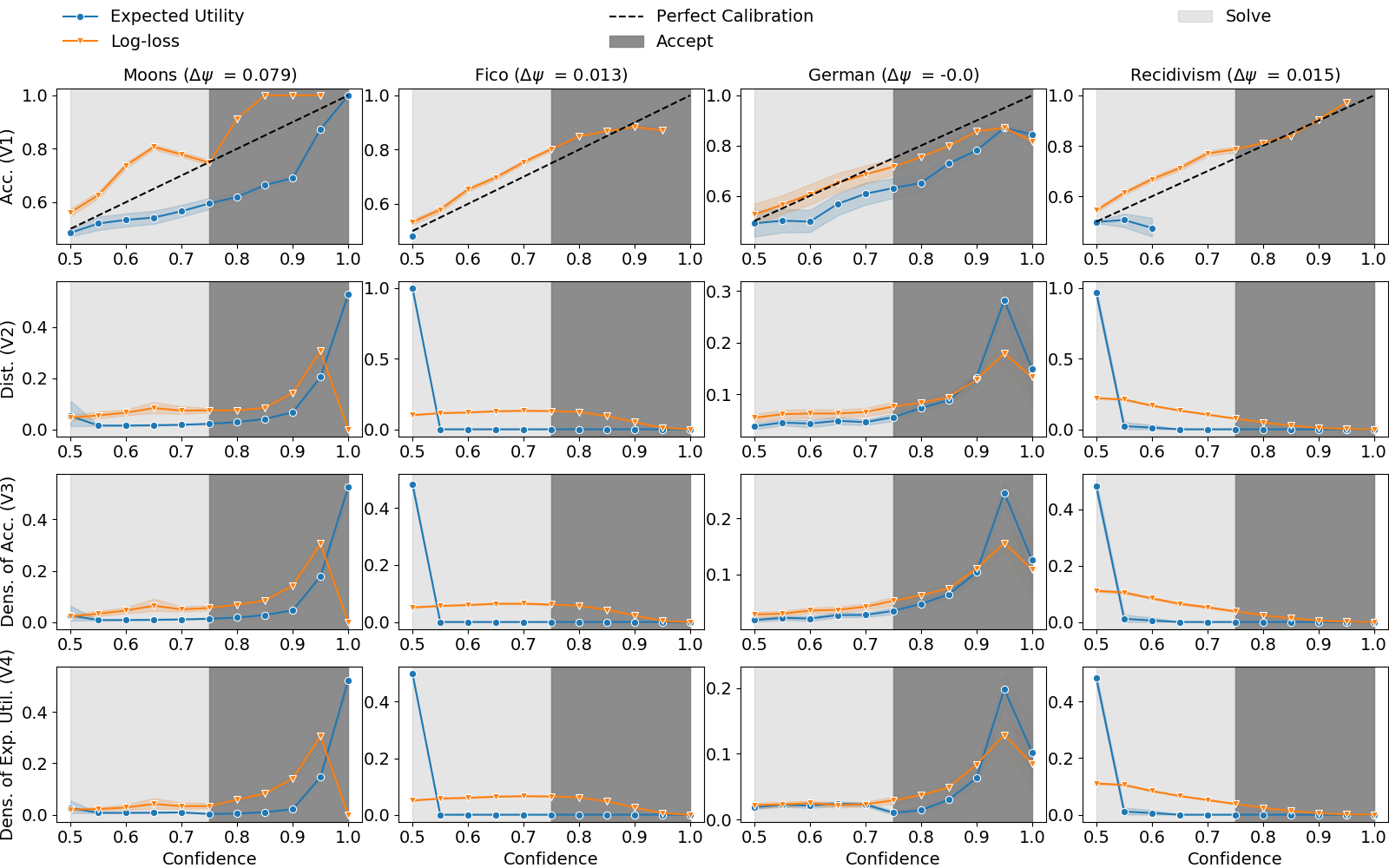}
    \caption{Behavior of new and old linear classifiers ($\mistakePenalty=1, \overrideCost=0.5, \humanAccuracy=1$) on rest of the datasets. On Moons and German we observe a similar shift in performance into the the \accept region. On Fico and Recidivism, as mentioned in the primary results before, the classifier learns that the expected utility of the most accurate classifier is less than that of the human performing the task by themselves, and hence, the classifier instead learns to be always uncertain. Note the new classifier still improve expected team utility, although for different reasons.}
\end{figure*}

\paragraph{Hyperparamters}
For each experiment setting, which is determined by combination of loss, dataset, classifier type, and task parameters, we selected best hyper-parameters using grid search, evaluating each parameter combination using 5-fold cross validation on 80\% of the available data, \ie, same amount of data that is available to training within a simulation: 
learning rate (1e-3, 1e-2, 1e-1, 1), L2 regularizer weight (1e-3, 1e-2, 1e-1), batch size (4, 8, 32), 
learning rate scheduler's decay factor (0.1, 0.9) and patience (2, 5, 10).


\begin{figure*}[t]
    \centering
    \includegraphics[width=0.8\textwidth]{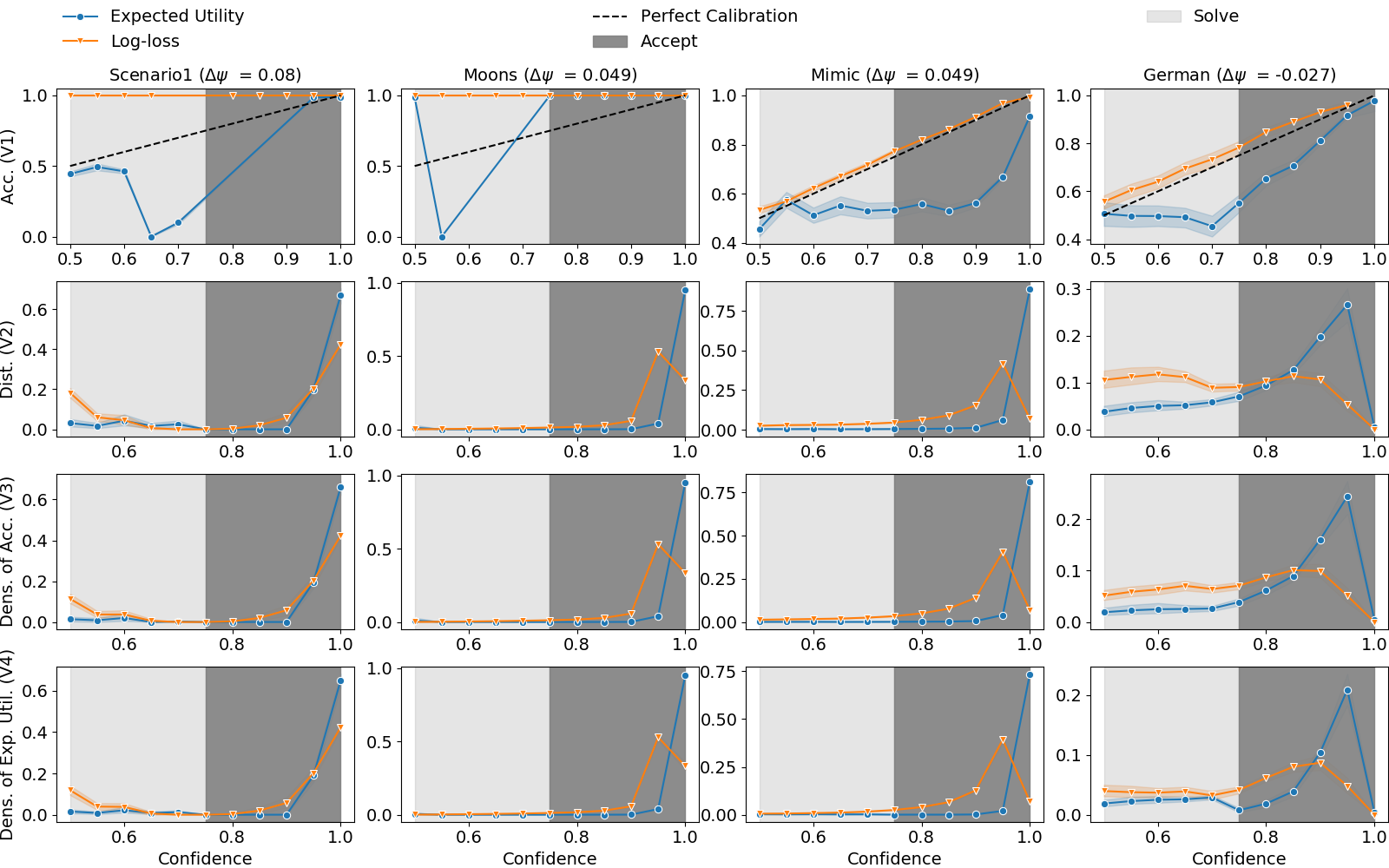}
    \caption{Differences in behavior of MLP on 4 out of 6 datasets ($\mistakePenalty=1, \overrideCost=0.5, \humanAccuracy=1$). On Scenario1, Moons, and Mimic we again see a similar shift in perform to the \accept region. Interestingly, on German, even though there was a decrease in mean expected utility we still see the retrained classifier attempts to improve performance in the \accept region. The decrease on German is perhaps explain by the decrease in utility from the low confidence regions (V4).}
\end{figure*}

\begin{figure}[t]
    \centering
    \includegraphics[width=\linewidth]{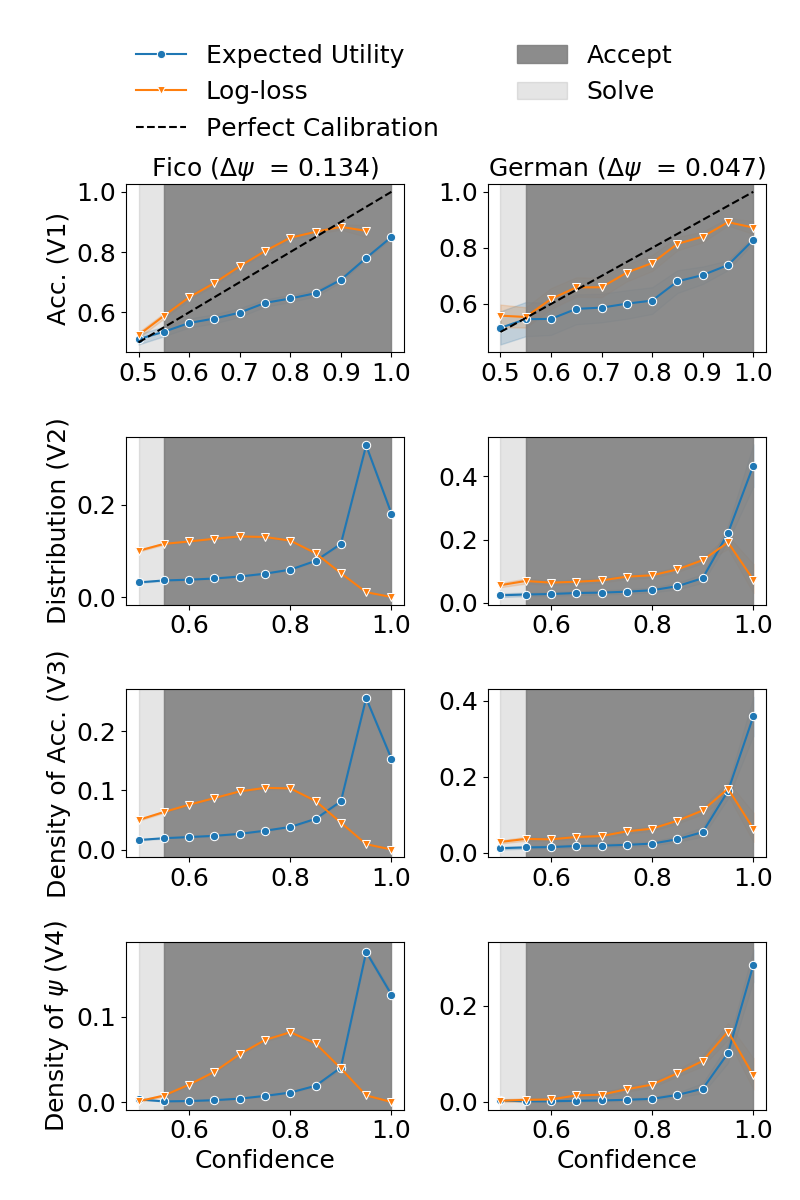}
    \caption{Differences in behavior of linear classifier when the human is not perfect ($\mistakePenalty=1, \overrideCost=0.5, \humanAccuracy=0.8$). On Fico and Recidivism, contrary to the case when human is perfect, the retrained classifier did not learn to be uncertain at all times. Instead, like for Mimic when human is perfect, the classifier improved performance on Fico and Recidivism in the \accept region. This is perhaps because as the human becomes less accurate the region in which the AI assistance is useful increases increasing the scope for the classifier to adjust and increase team's expected utility.}
\end{figure}

\end{document}